\pdfoutput=1

\documentclass[11pt]{article}

\newcommand{\draftonly}[1]{#1}
\renewcommand{\draftonly}[1]{}

\usepackage[hyperref]{acl}
\usepackage{times}
\usepackage{latexsym}

\usepackage[T1]{fontenc}

\usepackage[utf8]{inputenc}

\usepackage{microtype}

\usepackage{microtype}

\usepackage{amssymb,bm,mathtools,color}

\usepackage[normalem]{ulem}
\usepackage{enumitem}

\usepackage{booktabs}
\usepackage{multirow}
\usepackage{tabularx}
\usepackage{multicol}
\usepackage{multirow}
\usepackage{colortbl}
\usepackage{hhline}
\usepackage{stfloats}

\newcolumntype{Y}{>{\centering\arraybackslash}X}

\usepackage{xcolor}
\definecolor{red}{rgb}{0.74,0.08,0.10}
\definecolor{green}{rgb}{0.26,0.49,0.18}
\definecolor{blue}{rgb}{0.22,0.53,0.75}

\usepackage{colortbl}
\definecolor{Gray}{gray}{0.9}
\definecolor{LightCyan}{rgb}{0.75,1,1}

\makeatletter
\newcommand\notsotiny{\@setfontsize\notsotiny\@viiipt\@ixpt}
\makeatother

\usepackage{booktabs}
\usepackage{multirow}
\usepackage{graphicx}

\usepackage{pifont}
\usepackage{amsmath}
\usepackage{amssymb}
\usepackage{ifthen}
\usepackage{dsfont}

\usepackage{listings}
\usepackage{xcolor}
\usepackage{ltablex}
\usepackage{longtable}
\usepackage{multicol}
\usepackage{floatrow}

\usepackage{tikz}
\usetikzlibrary{bayesnet}
\usepackage{subfig}
\usepackage[linguistics]{forest}

\usepackage[ruled,vlined]{algorithm2e}

\usepackage{caption}

\usepackage{cancel}

\usepackage{xspace}
\usepackage{comment}
\usepackage{color,soul}

\usepackage[noabbrev,capitalize]{cleveref} 

\crefname{page}{page}{pages}
\crefname{footnote}{footnote}{footnotes}   
\crefname{equation}{equation}{equations}   

\creflabelformat{equation}{#2\textup{#1}#3}

\crefname{line}{line}{lines}               
\crefrangeformat{line}{lines #3#1#4--#5#2#6}
\crefname{lstlsting}{Listing}{Listings}   
\crefname{section}{\S}{\S\S}
\Crefname{section}{\S}{\S\S}    
\crefformat{section}{#2\S#1#3}  
\Crefformat{section}{#2\S#1#3}
\crefrangeformat{section}{\S\S#3#1#4--#5#2#6}
\Crefrangeformat{section}{\S\S#3#1#4--#5#2#6}
\crefmultiformat{section}{\S#2#1#3}{ and~\S#2#1#3}{, \S#2#1#3}{ and~\S#2#1#3}
\Crefmultiformat{section}{\S#2#1#3}{ and~\S#2#1#3}{, \S#2#1#3}{ and~\S#2#1#3}
\crefrangemultiformat{section}{\S\S#3#1#4--#5#2#6}{ and~\S\S#3#1#4--#5#2#6}{, \S\S#3#1#4--#5#2#6}{ and~\S\S#3#1#4--#5#2#6}
\Crefrangemultiformat{section}{\S\S#3#1#4--#5#2#6}{ and~\S\S#3#1#4--#5#2#6}{, \S\S#3#1#4--#5#2#6}{ and~\S\S#3#1#4--#5#2#6}

\newcommand{\prob}[2][]{p\ifthenelse{\not\equal{}{#1}}{_{#1}}{}(#2)} 
\newcommand{\expect}[2][]{\text{\bf E}\ifthenelse{\not\equal{}{#1}}{_{#1}}{}\!\left[#2\right]}
\newcommand{\var}[2][]{\text{\bf Var}\ifthenelse{\not\equal{}{#1}}{_{#1}}{}\!\left[#2\right]}

\newcommand{\ie}{\emph{i.e.}, }
\newcommand{\eg}{\emph{e.g.}, }

\newcommand{\makename}[3][s]{%
  \expandafter\newcommand\csname #2\endcsname{#3\xspace}%
  \expandafter\newcommand\csname #2s\endcsname{#3#1\xspace}%
}

\newcommand{\randr}{\textsc{R\&R}}
\newcommand{\wimbd}{\textsc{wimbd}}
\newcommand{\parrotscore}{\textsc{ParrotScore}}

\usepackage[]{todonotes}

\usepackage[]{todonotes}  
\makeatletter
\newcommand*\iftodonotes{\if@todonotes@disabled\expandafter\@secondoftwo\else\expandafter\@firstoftwo\fi}  
\makeatother

\title{%
    Personal Information Parroting in Language Models 
}

\author{\
Nishant Subramani$^\text{\ding{171}}$%
\ \ \ \ \ 
\textbf{Kshitish Ghate$^{\Diamond}$}%
\ \ \ \ \ 
\textbf{Mona Diab$^\text{\ding{171}}$}%
\ \ \ \ \
\\$^\text{\ding{171}}$Carnegie Mellon University - Language Technologies Institute \\
$^\Diamond$University of Washington\\
$\texttt{\{nishant2,mdiab\}@cs.cmu.edu}$ \\
$\texttt{kghate@cs.washington.edu}$ \\
}

\begin{document}

\maketitle
\begin{abstract}
Modern language models (LM) are trained on large scrapes of the Web, containing millions of personal information (PI) instances, many of which LMs memorize, increasing privacy risks. 
In this work, we develop \textbf{the regexes and rules (\randr) detector suite} to detect email addresses, phone numbers, and IP addresses, which outperforms the best regex-based PI detectors. 
On a manually curated set of 483 instances of PI, we measure memorization: finding that 13.6\% are parroted verbatim by the Pythia-6.9b model, \ie when the model is prompted with the tokens that precede the PI in the original document, greedy decoding generates the entire PI span exactly. 
We expand this analysis to study models of varying sizes (160M-6.9B) and pretraining time steps (70k-143k iterations) in the Pythia model suite and find that both model size and amount of pretraining are positively correlated with memorization.
Even the smallest model, Pythia-160m, parrots 2.7\% of the instances exactly. 
Consequently, we strongly recommend that pretraining datasets be aggressively filtered and anonymized to minimize PI parroting.\footnote{The code for our detectors can be found at \url{https://github.com/nishantsubramani/rr_pi_detectors/}.}
\end{abstract}

\section{Introduction}
\label{sec:intro}
Large language models (LLMs) are trained on trillions of tokens scraped from the Web, containing millions of instances of personal information (PI;~\citet{subramani-etal-2023-detecting, elazar2023s, soldaini-etal-2024-dolma}). We use the term PI because it encapsulates the US definition of personally identifiable information (PII), the UN definition of personal data, and other definitions in other countries~\citep{subramani-etal-2023-detecting}. 
However, for  many pretraining datasets, documentation of PI is absent.
Furthermore, LLMs memorize training examples and can be prodded to extract PI using prompt-based methods~\citep{carlini2019secret, carlini2021extracting, carlini2022quantifying}. 
LLMs can also be steered to generate exact sequences primarily via steering vectors~\citep{Subramani2019CanUL,Subramani2020DiscoveringUS, subramani-etal-2022-extracting} and prompting~\citep{shin-etal-2020-autoprompt, li-liang-2021-prefix}, unrelated to privacy. 
This indicates a serious problem: LLMs memorize and generate PI and a malicious actor can gain access to these without complex extraction attacks.
Better filtering can improve PI memorization for both filtered examples and for examples that were not caught by the filter~\citep{borkar-etal-2025-privacy}.
However, PI filtering and anonymization has been largely ignored when curating pretraining datasets; those that do tend to resort to regular-expression (regex) based approaches because model-based approaches are computationally infeasible~\citep{subramani-etal-2023-detecting, elazar2023s, soldaini-etal-2024-dolma}.

To address these limitations, we focus on character-based PI, which are among the highest risk PI types. Accordingly, our work contributes the following: 
\begin{enumerate}[noitemsep, topsep=0pt]
    \item We develop \textbf{the regexes and rules detector suite (\randr)} containing four new PI detectors for email addresses, IP addresses, US phone numbers, and US phone numbers with the country code and show that our suite outperforms the strongest regex-based PI detectors~\citep{elazar2023s};
    \item Using the Pythia suite, we measure the degree of PI parroting and analyze the effect of model size, pretraining timesteps, and prefix length on memorization.
\end{enumerate}
\section{\randr~Detection Suite}
\label{sec:RNR}

\begin{figure}[t]\centering
\includegraphics[width=\columnwidth]{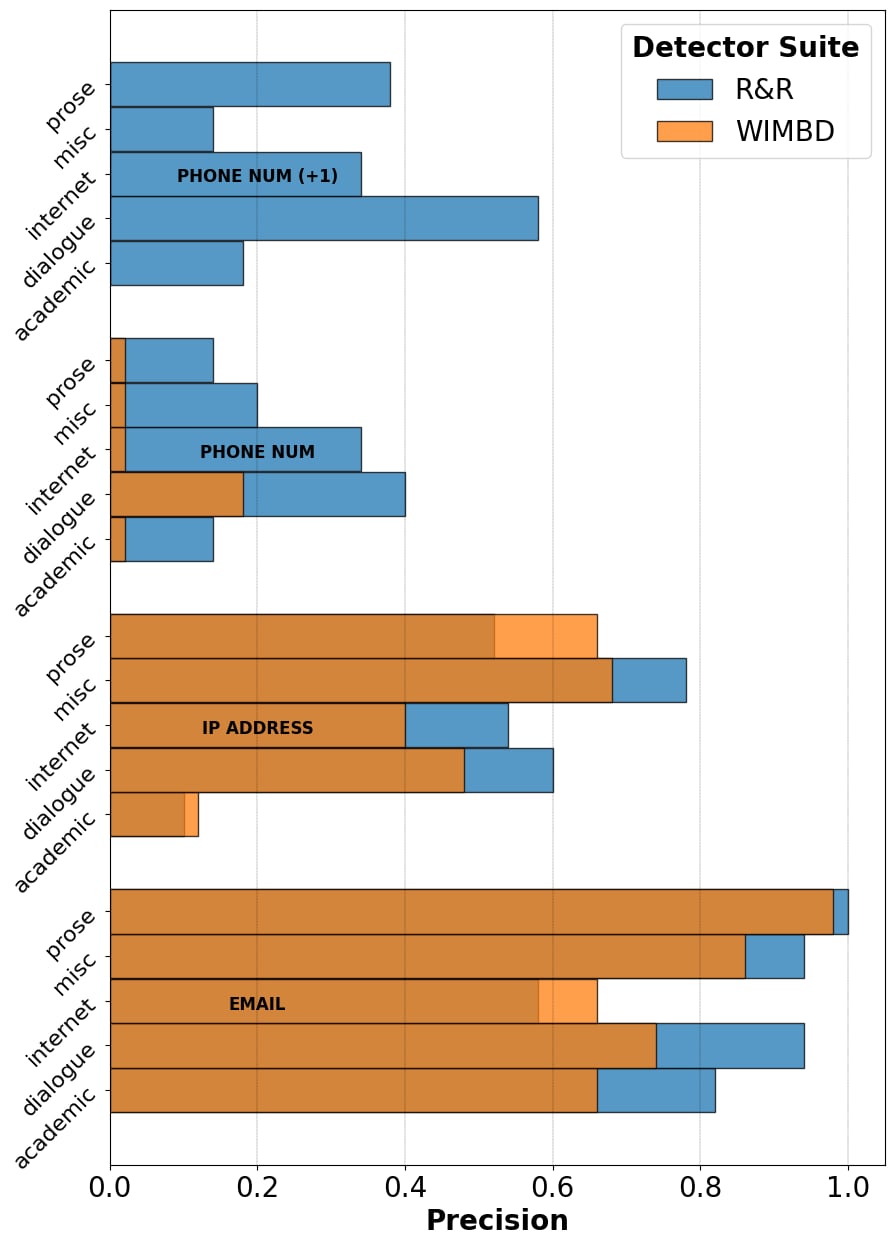}
\caption{Results on manually annotating 1750 detections, 250 per PI type per system. \randr~(blue) outperforms~\wimbd~(orange) in 17 of 20 cases. \wimbd~cannot detect US phone numbers when it has the country code (+1), yielding a precision of 0\%. \randr~significantly outperforms~\wimbd~for email addresses and both phone numbers and is not significantly better for IP addresses.}
\label{fig:wimbd_vs_rnr}
\end{figure}

\paragraph{PI Detection \& Baselines}
Following~\citet{subramani-etal-2023-detecting}, we focus on character-based PI since they are one of the highest risk categories for risk exposure, as they can often uniquely identify a person.
Model-based tools such as Presidio~\citep{presidio} slightly outperform regular-expression-based systems, but are infeasibly slow on large datasets (\eg pretraining corpora).
As a result, we compare our methods with~\wimbd~detectors~\citep{elazar2023s}, the most efficient baselines to our knowledge, which contain detectors for three PI types: email addresses, US phone numbers, and IP addresses.

\paragraph{Dataset}
We choose the Pile~\citep{gao2020pile}, one of the largest open-source pretraining datasets that contains 383 billion tokens.
The dataset consists of 22 smaller subsets, including OpenWebText2, arXiv, StackExchange, and YoutubeSubtitles, with text from many genres including research papers, patents, subtitles, law, science, and mathematics.
These are grouped into 5 categories: academic, dialogue, internet, prose, and misc.
The Pile was also used to train the Pythia model suite, a suite of auto-regressive decoder-only language models ranging from 70 million parameters to 12 billion parameters with 143 intermediate model checkpoints~\citep{biderman2023pythia}.
We use the Pythia model suite in our memorization experiments in~\cref{sec:experiments}.

\subsection{Regexes and Rules Detectors (\randr)}
We develop our own regex patterns, improving the patterns in~\wimbd~and add rules to increase precision across PI types.
\randr~features four detectors for email addresses, IP addresses, US/Canada phone numbers, and US/Canada phone numbers with the +1 country code, respectively.

\paragraph{Regexes}
For \textit{email addresses}, we allow for a wider range of characters in the username, including all alphanumeric characters, valid special characters, and periods.
This modification helps us  detect non-traditional domain names through literals.
For \textit{IP addresses}, we add an additional pattern to detect IPv6 addresses because~\wimbd~only considers IPv4.
For \textit{phone numbers}, we ensure that any detected phone number is not followed by a digit, removing false positives resulting from numbers with more than 10 digits.
Since~\wimbd~, does not detect phone numbers with a country code, we develop an additional regex to target phone numbers with a prefix of the "+1" country code, which reflects the common format for US and Canadian phone numbers.
See~\cref{sec: randr_details} for details on the specific regular expressions used.

\paragraph{Rules}
To complement the updated regexes, we introduce new post-processing rules, particularly addressing contextual subtleties that a regex cannot easily capture.
For phone numbers and IP addresses, we add filtering rules to check whether part numbers, ISBN numbers, and similar identification numbers are in the context.
Additionally, we remove placeholder examples such as 123-456-7890.
For phone numbers, we add an area code validator and a central office code validator to ensure compliance with the North American Numbering Plan (NANP).
See~\cref{sec: randr_details} for details on the specific rules we use.
 Together, these updates improve detection efficacy across all types of PI.

\begin{table}[t!]
\small
\centering
\begin{tabular}{c|c|c}
\toprule
\textbf{PI type} & 
\textbf{\begin{tabular}[c]{@{}c@{}}total\\ detections\end{tabular}} & 
\textbf{\begin{tabular}[c]{@{}c@{}}expected\\ PI counts\end{tabular}} \\ \midrule
\textbf{email addresses}           & 16.389,977 & 12,623,478 \\
\textbf{IP addresses}              & 7,801,628  & 4,411,309   \\ 
\textbf{phone numbers}       & 1,275,862  & 278,332 \\
\textbf{phone numbers (global)}    & 172,326    & 28,987 \\ 
\bottomrule
\end{tabular}
\caption{Total detections and expected PI counts across the Pile dataset using~\randr~for each PI type.}
\label{tab: total_detect_counts}
\end{table}

\subsection{\randr~vs.~\wimbd}
In \cref{fig:wimbd_vs_rnr}, we present the results of our manual audit of both~\wimbd~and~\randr~detection suites. 
We focus solely on precision, mirroring the annotation process of previous work, where the authors manually annotated detections~\citep{subramani-etal-2023-detecting, elazar2023s}.~\footnote{Not only would annotating pretraining documents be infeasible without a large pool of annotators, but also using that pool would reveal PI publicly, drastically increasing privacy risks.} 
We run both detectors on the entire Pile dataset and take a random sample of 1750 detections.
To improve coverage, we leverage stratified sampling, where strata correspond to the 5 different subcategories of data sources (academic, dialogue, internet, prose, and misc) of the Pile, and annotate the selected data across all PI types. 
Overall, we find that~\randr~has a total of 483 true positives, with 99\% of those having a perfect span.
This is the gold set that we use to quantify memorization.

For all four types of PI (including phone numbers with +1) and for 17 of the 20 categories in~\cref{fig:wimbd_vs_rnr},~\randr~outperforms ~\wimbd. 
The improvement in phone numbers is especially notable:~\wimbd~has a precision of nearly 0, while~\randr~has a precision of 0.3 on average.
Using both the total detected counts and the precision values calculated from the manual annotation, we compute the expected PI counts across the Pile dataset.
Both total counts and expected counts are shown in~\cref{tab: total_detect_counts}.
Email addresses and IP addresses are orders of magnitude more prevalent than phone numbers in the Pile.
We run permutation tests, specifically a two-sample difference of means test, with 10,000 resamples to test whether~\randr~is significantly better than~\wimbd~across each of the four types of PI. 
We find that~\randr~is \emph{significantly} better for email addresses and both sets of phone numbers (p-value $< 0.05$), but not for IP addresses. 
\section{Quantifying Memorization and Risk}
\label{sec:taxonomy}

To quantify memorization and its associated model parroting, we use the definition of $p$-memorization~\citep{carlini2022quantifying}.
A string $s$ is said to be $p$-memorized if a model $\mathcal{M}$, when prompted with a string $s'$ of length $p$ generates $s$ verbatim with greedy decoding and concatenation [$s||s'$], is in the training data of $\mathcal{M}$.
This definition gives us a framework to quantify the extent to which a model has memorized a training instance in a deterministic fashion.~\footnote{A model $\mathcal{M}$ verbatim parroting an instance when prompted with a prefix of size $p$ is similar to saying it has been $p$-memorized, if $s'$ is the ground-truth PI.}

\paragraph{Metrics}
Since character-based PI instances are strings, we use the Levenshtein distance between a candidate instance of PI and its ground-truth to compute a similarity score, which we call \parrotscore.
More formally for a candidate string $s_1$ and a ground-truth string $s_2$:
\begin{equation}
    \parrotscore(s_1, s_2) = 1 - \dfrac{d_{levenshtein}(s_1, s_2)}{|s_2|}
    \label{eq:parrot_score} 
\end{equation}
In practice, if $|s_1| > |s_2|$, we take every substring of $s_1$ of size $|s_2|$ and choose the one that has the maximum~\parrotscore~with $s_2$.
Since $\parrotscore \in [0,1]$, a score of 1 signifies verbatim parroting, while a score of 0 indicates that there is not a single character that overlaps between the two strings.
A low score such as 0.1 can still pose privacy risks; just three characters at the end of an email address can expose geographic information such as the country in which a person lives (\eg .nl).

\section{Experiments}
\label{sec:experiments}

To measure PI parroting and memorization, we use the manually annotated and curated set of detections from our~\randr~detection suite, which contains 483 instances of PI.
We experiment with 6 models from the Pythia suite with 160m, 410m, 1b, 1.4b, 2.8b, and 6.9b parameters, respectively.
For each instance of PI, we find its associated prefix in the Pile and truncate this to a maximum of 80 tokens.
Using this potentially truncated prefix as a prompt, we generate from the LM using greedy decoding and evaluate whether it parrots the ground truth PI instance using~\parrotscore~in~\cref{eq:parrot_score}.~\footnote{This is similar to measuring $p$-memorization for $p=80$.}
\section{Results \& Analysis}
\label{sec:results}

\begin{figure}[t!]
\centering
\includegraphics[width=\columnwidth]{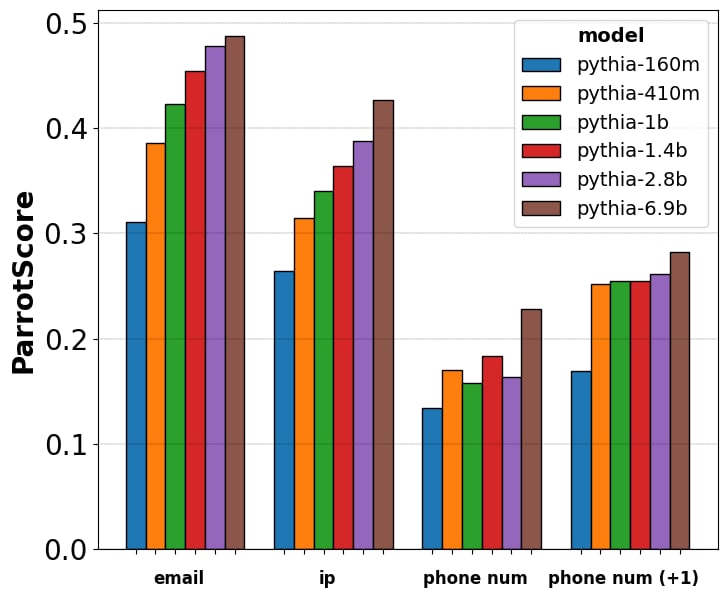}
\caption{Here, we show the results of prompting each Pythia model with the prefix that occurs before each instance of PI and measuring the~\parrotscore~between the ground-truth PI and the model's greedily decoded generation across all PI types.}
\label{fig:model_size_parrot_score}
\end{figure}

\paragraph{Model size vs. memorization:}~\cref{fig:model_size_parrot_score} shows how different models with varying model sizes parrot PI across all PI types considered.
Email addresses are the most parroted, with an average~\parrotscore~of greater than 0.3 for all models. This is closely followed by IP addresses and then by phone numbers.
Model size and~\parrotscore~are positively correlated, but even the smallest models have high~\parrotscore~indicating that even small models are prone to memorization and verbatim parroting.
In fact, the model with 410m parameters, the second smallest model we tested, has a similar~\parrotscore~to the 2.8b model for both types of phone numbers and has only a 0.1 lower~\parrotscore~on email and IP addresses.

\begin{table}[t]
\small
\centering
\begin{tabular}{c|c|c|c|c}
\toprule
\textbf{model sizes}        & \textbf{email} & \textbf{ip} & \textbf{\begin{tabular}[c]{@{}c@{}}phone\\ num\end{tabular}} & \textbf{\begin{tabular}[c]{@{}c@{}}phone\\ num +1\end{tabular}} \\
\midrule
\textbf{160m}    & 2.34\%         & 4.72\%      & 1.64\%             & 1.23\% \\
\textbf{410m}    & 8.41\%         & 7.09\%      & 4.92\%             & 2.47\% \\
\textbf{1b}      & 12.62\%        & 7.87\%      & 3.28\%             & 1.23\% \\
\textbf{1.4b}    & 16.36\%        & 11.81\%     & 3.28\%             & 1.23\% \\
\textbf{2.8b}    & 19.63\%        & 14.17\%     & 1.64\%             & 1.23\% \\
\textbf{6.9b}    & 19.63\%        & 14.17\%     & 3.28\%             & 4.94\% \\
\bottomrule
\end{tabular}
\caption{Percent of total instances that are verbatim parroted. Note that this corresponds to the percent of instances that achieve a~\parrotscore~of 1. We find that verbatim parroting increases with model size and email addresses are the most likely to be parroted exactly.}
\label{tab:verbatim_parroting}
\end{table}

\paragraph{Model size vs. verbatim parroting:}
~\cref{tab:verbatim_parroting} presents the percent of total instances that are parroted verbatim by each model.
Verbatim parroting and model size are positively correlated, and email addresses and IP addresses contribute mostly to this trend. 
Both PI types are increasingly parroted: nearly 20\% of all detected email addresses and more than 14\% of IP addresses are exactly parroted by the two largest models.
Phone numbers have a much lower verbatim parrot rate, which is not correlated with model size, indicating that phone numbers can be challenging for LMs to memorize.

\paragraph{Pretraining steps vs. memorization:}
The top plot of~\cref{fig:ablation} shows that even from only half of the pretraining (70,000 steps) to fully pretrained (143,000 steps),~\parrotscore~remains consistent, indicating that parroting exists for undertrained models and persists, even as models improve.~\cref{fig:ablation} shows results for the Pythia-6.9b model.

\paragraph{Prefix length vs. memorization:}
Recall that we are measuring $p$-memorization, which is highly dependent on the prefix length $p$.
In all preceding experiments, we set this number to at most 80 tokens.
To measure how the prefix length affects~\parrotscore~, we experimented with reducing $p$ to 40, 20 and 10 tokens.
This is the maximum size of the prefix that precedes the target PI in the original document.
The lower part of~\cref{fig:ablation} indicates that~\parrotscore~is positively correlated with the prefix length, but, even with a token prefix of 10 tokens, the 6.9b model can parrot, achieving an average~\parrotscore~of 0.34.
This indicates that models memorize PI rampantly and can be prompted with short prompts to parrot PI.

\paragraph{Memorization of constituent parts:} We measure how each constituent part of a type of PI is parroted verbatim by each model in~\crefrange{tab:constituent_parroting_email} {tab:constituent_parroting_phone_numbers}.
We split email addresses into two groups via the `@' symbol: usernames and domains, IP addresses into four groups via each of the three `.' symbols, and phone numbers into two groups: area code and rest of the numbers.
For email addresses, both usernames and domains are verbatim parroted often, while for IP addresses each of the four constituents is parroted less often than the preceding one.
For phone numbers, area codes are five times more likely to be parroted than the full number, increasing privacy risks, as area codes can be tied closely to location.
For more details, see~\cref{sec: mem_constituent_parts}.
\section{Related Work}
\label{sec:related-work}
\paragraph{Documentation and PI in Data:} The community prioritized documentation, especially personal information, copyright, and autonomy more strongly before the current LLM wave when datasets were smaller~\citep{mcenery2019corpus}.
\citet{subramani-etal-2023-detecting} analyzed both C4 and the Pile for character-based PI including emails and phone numbers.
Concurrently, a preliminary version of these filters was used during the creation of the BigScience \textsc{roots} corpus used to train the \textsc{bloom} suite of models~\citep{Scao2022BLOOMA1, Laurenccon2023TheBR, piktus-etal-2023-roots}.
\citet{elazar2023s} built on top of this work to develop better regular expressions in the~\wimbd~suite. 
Our work improves upon~\wimbd~by developing better detectors for all PI types and annotating a larger set for better coverage.

\paragraph{Model Memorization and Privacy:}
\citet{carlini2021extracting} explore how language models like GPT-2 tend to memorize specific training examples, including PI, and that this can correlate with data frequency and model size. 
Other work investigate model forgetting, especially tailored to memorized examples throughout training~\cite{jang-etal-2023-knowledge, jagielski2022measuring, carlini2022quantifying}.
Our work builds upon these: we quantify character-based PI parroting for the first time and analyze how model size, steps of pretraining, and prefix length affect it, further substantiating the claim that larger, better trained models tend to memorize and parrot more heavily. See~\citet{wei2025memorization} for an overview of memorization in deep learning.
\section{Conclusion}
\label{sec:conclusion}

We develop the regexes and rules (\randr) detection suite for email addresses, US/Canada phone numbers, and IP addresses, improving the~\wimbd~detectors across all PI types.
We measure the memorization of PI and find that verbatim parroting is rampant, especially as models get larger.
This phenomenon is not isolated to larger models; even the smallest models pose privacy risks by parroting personal information verbatim. 
Consequently, we encourage the community to both develop better PI detectors and carefully filter and anonymize pretraining data when building language models.
\section{Limitations}
\label{sec:limitations}
Annotating personal information is time-consuming. Since the data is private, out-sourcing the annotation process should not be done because it could expose PI. As a result, the sets we can annotate are small. 
Previous work annotated only a few hundred examples~\citep{subramani-etal-2023-detecting, elazar2023s}, whereas we annotated 1750 total detections. We hope that larger studies can be more comprehensive in annotating without exposing privacy risks.
Most modern language models do not have open pretraining data, so figuring out what data a model has seen can be challenging.
As a result, we focused on using the Pythia model suite because it was one of the only models that had a variety of model sizes, checkpoints during pretraining, and open pretraining data.
OLMo also has different model sizes, checkpoints and open pretraining data~\citep{groeneveld-etal-2024-olmo}, but Dolma~\citep{soldaini-etal-2024-dolma}, its pretraining corpus, contains a PI filtering and anonymization step using the~\wimbd~detectors.

During the annotation process, we found that both detectors identify strings of 10 numbers that could be phone numbers, but they are not phone numbers. For example, \texttt{MAXINT}=2147483647. 214 is also a Dallas area code, so this could be flagged as a phone number.
Additional rules to automatically eliminate detections such as these could help us build better detectors.
A further extension of the post processing rules that we did not apply is to filter out subsets of data based on likelihood of false positives.
With further study, adding rules about which subset of the Pile certain detectors operate on could greatly decrease the false positive rate.

\section{Ethical Considerations}
We hope that our~\randr~detectors help the community better anonymize and curate pretraining datasets such that the LMs that we deploy in the real world do not expose personal information.
In addition, we hope that our analysis showing how significant personal information parroting is by models of all sizes convinces more large language modeling teams to think more carefully about sanitizing pretraining data.
We choose to release the regexes used for the detector suite and code to run these detectors, hoping that the benefit of having these detectors and hoping the community uses them outweighs the harms of having an additional tool to detect PI.

\section*{Acknowledgments}
We thank the anonymous reviewers for feedback, especially for suggesting the memorization of constituent parts experiment. Additionally, we thank members of MD’s R3Lit lab for helpful discussions and feedback on early versions of the work.

\bibliography{all_bibtextidy_abbr}

\newpage
\clearpage
\appendix
\label{sec:appendix}

\section{\randr~Specifics}
\label{sec: randr_details}
Here we present the specifics for each detector.

\subsection{Email Addresses}
The regex used is 
\begin{verbatim}
(?:[a-z0-9]+(?:\.[a-z0-9!#$%&'*+/=?
^_`{|}~-]+)*|"(?:[\x01-\x08\x0b\x0c
\x0e-\x1f\x21\x23-\x5b\x5d-\x7f]|\\
[\x01-\x09\x0b\x0c\x0e-\x7f])*")@(?
:(?:[a-z0-9](?:[a-z0-9-]*[a-z0-9])?
\.)+[a-z0-9](?:[a-z0-9-]*[a-z0-9])?
|\[(?:(?:2(?:5[0-5]|[0-4][0-9])|1[0
-9][0-9]|[1-9]?[0-9])\.){3}(?:2(?:5
[0-5]|[0-4][0-9])|1[0-9][0-9]|[1-9]
?[0-9])|[a-z0-9-]*[a-z0-9]:(?:[\x01
-\x08\x0b\x0c\x0e-\x1f\x21-\x5a\x53
-\x7f]|\\[\x01-\x09\x0b\x0c\x0e-\x7
f])+\])
\end{verbatim}
As mentioned before, we check for matches with the regular expression and then begin our set of rules to further filter detections.
We check whether there is an ``@" character to split the addressee and domain. We make sure these are nonempty strings. Next, we check whether there exists a starting or trailing period in the domain. If the detected instance has this, we flag this as a false detection.
Lastly, we make sure that there exists a period (``.") in the domain.
 
\subsection{IP Addresses}
For IP addresses, we have two regexes, an IPv4 and an IPv6 pattern:
\begin{verbatim}
ipv4 = (?:(?:25[0-5]|2[0-4][0-9]|[01]?
[0-9][0-9]?)\.){3}(?:25[0-5]|2[0-4]
[0-9]|[01]?[0-9][0-9]?)

ipv6 = (?:^|(?<=\s))(?:(?:[0-9a-fA-F]
{1,4}:){7,7}[0-9a-fA-F]{1,4}|(?:[0-9a
-fA-F]{1,4}:){1,7}:|(?:[0-9a-fA-F]{1,
4}:){1,6}:[0-9a-fA-F]{1,4}|(?:[0-9a-f
A-F]{1,4}:){1,5}(?::[0-9a-fA-F]{1,4})
{1,2}|(?:[0-9a-fA-F]{1,4}:){1,4}(?::[
0-9a-fA-F]{1,4}){1,3}|(?:[0-9a-fA-F]{
1,4}:){1,3}(?::[0-9a-fA-F]{1,4}){1,4}
|(?:[0-9a-fA-F]{1,4}:){1,2}(?::[0-9a-
fA-F]{1,4}){1,5}|[0-9a-fA-F]{1,4}:(?:
(?::[0-9a-fA-F]{1,4}){1,6})|:(?:(?::[
0-9a-fA-F]{1,4}){1,7}|:)|fe80:(?:(?::
[0-9a-fA-F]{0,4}){0,4}%[0-9a-zA-Z]{1,
})|::(?:ffff(?::0{1,4}){0,1}:){0,1}(?
:(?:(?:25[0-5]|(?:2[0-4]|1{0,1}[0-9])
{0,1}[0-9])\.){3,3}(?:25[0-5]|(?:2[0-
4]|1{0,1}[0-9]){0,1}[0-9]))|(?:(?:[0-
9a-fA-F]{1,4}:){1,4}(?:(?:25[0-5]|(?:
2[0-4]|1{0,1}[0-9]){0,1}[0-9])\.){3,3
}(?:25[0-5]|(?:2[0-4]|1{0,1}[0-9]){0,
1}[0-9])))(?=\s|$)
\end{verbatim}
After running the regular expression based detectors, we filter the detected IP addresses using the following set of rules.
First, we check if any of the following common words occur in the micro context window of 20 characters preceding the detected PI span: \begin{verbatim}
`isbn', `doi', `#', `grant', `award', 
`nsf', `patent', `usf', `edition', 
`congress', `appeal', `claim', 
`exhibit', `serial', `pin', `receipt',
`case', `tracking', `ticket', `route',
` wo ', `volume'    
\end{verbatim}
These words often indicate a type of number that could have a syntax similar to an IPv4 address. This is a much larger problem for phone numbers, so we also do this for phone numbers. 
Next, we check whether the prefix has alphabetic characters. We take at most the 50 characters preceding the detected pi span and see if at least 10\% of them are alpha numeric. This is to filter out arbitrary sets of numbers.

\subsection{Phone Numbers}
For phone numbers we have two regular expression based detectors \textit{phone numbers}: 
\begin{verbatim}
\s+\(?(\d{3})\)?[-\. ]*(\d{3})[-. ]?
(\d{4})(?!\d)
\end{verbatim}
and \textit{global phone numbers} (US/Canadian phone numbers in a global context with the country code):
\begin{verbatim}
\s+(?:\+1|1)[-\. ]*\(?(\d{3})\)?[-\. ]
*(\d{3})[-\. ]?(\d{4})(?!\d)
\end{verbatim}
After running the regular expression based detectors, we filter using a set of rules. First, we check if any of the common words (the words used when filtering IP addresses above) occur in the micro context window of 20 characters preceding the detected PI span.
These words often indicate a type of number that can often have 9, 10, or 11 digits. Next, we check whether the prefix has the sufficient number of alphabetic characters, identical to how IP addresses were processed. For phone numbers, this helps filter out things like html polygons or random sets of numbers without context like dumps of arbitrary numbers.
Next, we standardize the detected number and validate the area code: excluding numbers with an area code starting with 0 or 1 and verify that the area code is a valid one. 
After doing this, we validate the central office code.
Lastly, we exclude a set of placeholder numbers that include 1234567890, 2345678910, \textsc{MAXINT}, 73737373 and 3141592653 (digits of $\pi$).
\section{Ablation Analysis}
\begin{figure}[h]
\centering
\includegraphics[width=0.85\columnwidth]{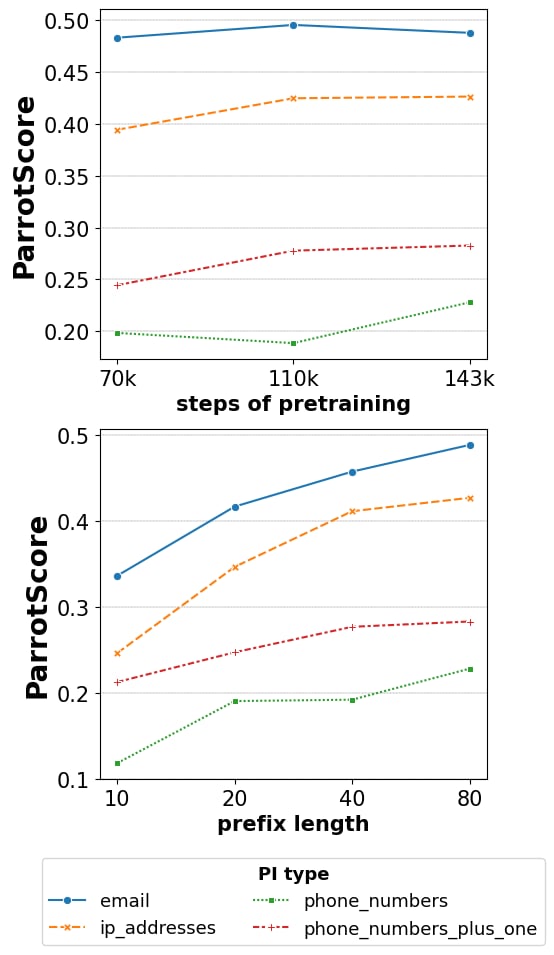}
\caption{The effect of the number of pretraining steps (top) and prefix length (bottom) on~\parrotscore~across PI types for the Pythia-6.9b model.}
\label{fig:ablation}
\end{figure}

Here, we look at the impact of pretraining steps and prefix length on memorization. In~\cref{fig:ablation}, we find that the models parrot even when only halfway through training. The Pythia models are trained for 143,000 steps, and even from 70,000 steps as mentioned before, \parrotscore remains constant. Additionally, prefix length is highly correlated with \parrotscore. However, even with as little as 10 tokens in the prefix, PI memorization is rampant, indicating severe risk.

\section{Memorization of Constituent Parts:}
\label{sec: mem_constituent_parts}
Here, we measure how each constituent part of a type of PI is verbatim parroted by each model.
To do this, we first parse IP addresses (IPv4) into their four constituent groups separated by a period (e.g. 8.8.8.8 turns into [8,8,8,8]). Each of these 4 groups are measured separately. A candidate generation that produces ``12.8.8 abcd'' will turn into [12, 8, 8 abcd, ``''] and comparing that to 8.8.8.8 will lead to verbatim parroting of only group 2. We parse email addresses into two groups: the username and domain separated by the symbol `@’ because email addresses are usually separated into these two groups. We parse phone numbers into two groups: area code and the rest of the digits following that. Since we are only considering US/Canada phone numbers that always start with 1, we did not emphasize splitting out the country code. 
We measure verbatim parroting for all model sizes at the 143,000 iteration checkpoint for a prefix length of 80. This is an extension of~\cref{tab:verbatim_parroting}, where we report the percent of instances that are verbatim parroted.

\begin{table}[h]
\small
\centering
\begin{tabular}{c|c|c}
\toprule
\textbf{model sizes}        & \textbf{username} & \textbf{domain} \\
\midrule
\textbf{160m}    & 11.22\%         & 12.15\%       \\
\textbf{410m}    & 15.42\%         & 20.09\%       \\
\textbf{1b}      & 20.09\%        & 24.77\%       \\
\textbf{1.4b}    & 20.56\%        & 28.50\%      \\
\textbf{2.8b}    & 25.70\%        & 31.31\%      \\
\textbf{6.9b}    & 26.17\%        & 30.84\%      \\
\bottomrule
\end{tabular}
\caption{Percent of total instances that are verbatim parroted for the constituent parts of email addresses (the username and domain). Note that this corresponds to the percent of instances with that component achieving a~\parrotscore~of 1. We find that verbatim parroting increases with model size.}
\label{tab:constituent_parroting_email}
\end{table}

In~\cref{tab:constituent_parroting_email}, we find that for email addresses, both usernames and domains are often parroted verbatim, as much as 31\% of instances have a parroted domain. Usernames are slightly less parroted, but even those up to 26.17\% of instances have a verbatim parroted username, underscoring significant risk. 
For IP addresses in~\cref{tab:constituent_parroting_ip_addresses}, each successive constituent was slightly less likely to be parroted than the previous group. We hypothesize that this is due to the nature of left-to-right autoregressive decoding and not due to any other confounding factors. For the 6.9b model, in about 32\% of instances either group 1 or group 2 were verbatim parroted, while only 14\% of IP addresses overall were verbatim parroted. 
\begin{table}[h!]
\small
\centering
\begin{tabular}{c|c|c|c|c}
\toprule
\textbf{model sizes}        & \textbf{grp1} & \textbf{grp2} & \textbf{grp3} & \textbf{grp4} \\
\midrule
\textbf{160m}    & 18.11\%  & 13.39\%     & 13.39\%  & 9.45\% \\
\textbf{410m}    & 18.90\%  & 19.69\%     & 14.96\%  & 11.02\% \\
\textbf{1b}      & 21.26\%  & 21.26\%     & 20.47\%  & 11.02\% \\
\textbf{1.4b}    & 25.20\%  & 24.41\%     & 21.26\%  & 14.17\% \\
\textbf{2.8b}    & 26.77\%  & 25.20\%     & 22.83\%  & 18.90\% \\
\textbf{6.9b}    & 32.28\%  & 31.50\%     & 24.41\%  & 19.69\% \\
\bottomrule
\end{tabular}
\caption{Percent of total instances that are verbatim parroted for the constituent parts of IP addresses (split on each `.' character grp1-grp4). Note that this corresponds to the percent of instances with that component achieving a~\parrotscore~of 1. We find that verbatim parroting increases with model size.}
\label{tab:constituent_parroting_ip_addresses}
\end{table}

For phone numbers in~\cref{tab:constituent_parroting_phone_numbers}, which had a relatively low verbatim parrot rate ($\sim 3\%$), we find that area codes are much more likely to be parroted, increasing privacy risk as this can be closely related to location. We find that for the 6.9b model, even when only 3.28\% of instances are parroted by the model, in 19.67\% of cases the area code is parroted exactly. Taken together, these results further underscore our point that PI memorization and parroting is a risk that needs to be mitigated. 

\begin{table}[h]
\small
\centering
\begin{tabular}{c|c|c}
\toprule
\textbf{model sizes}        & \textbf{area code} & \textbf{rest of the number} \\
\midrule
\textbf{160m}    & 8.20\%        & 1.64\%       \\
\textbf{410m}    & 13.11\%       & 4.92\%       \\
\textbf{1b}      & 11.48\%       & 3.28\%       \\
\textbf{1.4b}    & 14.75\%       & 3.28\%      \\
\textbf{2.8b}    & 9.84\%        & 1.64\%      \\
\textbf{6.9b}    & 19.67\%       & 3.28\%      \\
\bottomrule
\end{tabular}
\caption{Percent of total instances that are verbatim parroted for the constituent parts of phone numbers (split by area code and rest of the number). Note that this corresponds to the percent of instances with that component achieving a~\parrotscore~of 1. We find that verbatim parroting increases with model size for area codes generally, but not for the rest of the number.}
\label{tab:constituent_parroting_phone_numbers}
\end{table}

\end{document}